\documentclass{article}
\usepackage{amsmath,booktabs,tikz,hyperref}
\usepackage[preprint]{spconf}

\title{A Novel Image Similarity Metric for Scene Composition Structure}

\name{Md Redwanul Haque$^{\star}$ \qquad Manzur Murshed$^{\star}$ \qquad Manoranjan Paul$^{\dagger}$ \qquad Tsz-Kwan Lee$^{\star}$}
\address{$^{\star}$ School of Information Technology, Deakin University, Burwood, Victoria, Australia \\
$^{\dagger}$ School of Computing, Mathematics and Engineering, Charles Sturt University, Bathurst, NSW, Australia}

\newcommand{\verticaltoappear}[1]{
    \begin{tikzpicture}[remember picture,overlay]
    \node[anchor=south, xshift=-12mm, rotate=-90] at (current page.east) {#1};
    \end{tikzpicture}
}
\begin{document}

\maketitle
\copyrightnotice{\parbox{\textwidth}{\footnotesize © 2025 IEEE. Personal use of this material is permitted. Permission from IEEE must be obtained for all other uses, in any current or future media, including reprinting/republishing this material for advertising or promotional purposes, creating new collective works, for resale or redistribution to servers or lists, or reuse of any copyrighted component of this work in other works.}}
\verticaltoappear{\footnotesize \it{2025 IEEE ICIPW, 14-17 Sept., 2025, Anchorage, Alaska, USA}}

\begin{abstract}
The rapid advancement of generative AI models necessitates novel methods for evaluating image quality that extend beyond human perception. A critical concern for these models is the preservation of an image's underlying Scene Composition Structure (SCS), which defines the geometric relationships among objects and the background, their relative positions, sizes, orientations, etc. Maintaining SCS integrity is paramount for ensuring faithful and structurally accurate GenAI outputs. Traditional image similarity metrics often fall short in assessing SCS. Pixel-level approaches are overly sensitive to minor visual noise, while perception-based metrics prioritize human aesthetic appeal, neither adequately capturing structural fidelity. Furthermore, recent neural-network-based metrics introduce training overheads and potential generalization issues. We introduce the SCS Similarity Index Measure (SCSSIM), a novel, analytical, and training-free metric that quantifies SCS preservation by exploiting statistical measures derived from the Cuboidal hierarchical partitioning of images, robustly capturing non-object-based structural relationships. Our experiments demonstrate SCSSIM's high invariance to non-compositional distortions, accurately reflecting unchanged SCS. Conversely, it shows a strong monotonic decrease for compositional distortions, precisely indicating when SCS has been altered. Compared to existing metrics, SCSSIM exhibits superior properties for structural evaluation, making it an invaluable tool for developing and evaluating generative models, ensuring the integrity of scene composition.
\end{abstract}

\begin{keywords}
Image analysis, Image similarity metric, Scene composition structure, Hierarchical image partitioning, Generative AI
\end{keywords}

\section{Introduction}
\label{sec:intro}

Recent advancements in deep learning have fueled the rapid rise of generative AI (GenAI) \cite{rombach_high-resolution_2022}. Initially, GenAI models primarily focused on synthesising realistic images optimised for human visual perception. However, the utility of GenAI now extends to diverse applications where its outputs are processed or analysed automatically. For instance, in learning-based image/video coding, where generated images are used for downstream machine tasks \cite{yang_vcm_2024,rajin_forward_2022}. In such contexts, the accurate reproduction of scene content often takes precedence over perceptual realism for humans. For generated images to be reliably used in such analytical pipelines, akin to original images, they must preserve key object attributes like shape, size, orientation, and their spatial relationships within the scene and to each other \cite{yang_modeling_2022}. We collectively term these crucial properties the \textit{Scene Composition Structure} (SCS).

Currently, efforts to generate reproducible images with GenAI often rely on conditioning models with object-based side information like semantic segmentations, edge maps, scene graphs, or bounding-box layouts \cite{yang_modeling_2022, zhang_controlnet_2023, bansal_universal_2023}. However, these object-based, data-driven approaches face limitations, as their performance can degrade significantly with unknown or out-of-distribution objects. Furthermore, such side information is often not readily compressible, rendering it suboptimal for compression-focused applications. Future advancements may therefore require GenAI models for reproducible image generation to be conditioned directly on SCS, leveraging non-object-based side information to overcome these challenges.

However, developing such models also requires metrics capable of evaluating their success in preserving SCS. Commonly used image similarity metrics such as PSNR, SSIM \cite{wang_image_2004}, MS-SSIM \cite{wang_multiscale_2003}, FID \cite{heusel_gans_2017}, LPIPS \cite{zhang_unreasonable_2018}, and CLIP Score \cite{hessel_clipscore_2021} are ill-suited for this task. These metrics were primarily designed to quantify pixel-level fidelity (exact reproducibility) or to align with human perceptual judgements of realism (realistic synthesis), rather than assessing the preservation of SCS (pseudo-reproducibility). Consequently, there is a clear need for a novel image similarity metric that evaluates SCS by leveraging non-object-based structures within the scene without requiring model training, a dependency present in some contemporary learned metrics \cite{heusel_gans_2017, zhang_unreasonable_2018, hessel_clipscore_2021}.

In this paper, we have proposed a new image similarity metric for SCS, after first formally defining SCS in images and then showing the ineffectiveness of the existing similarity metrics in assessing SCS. The following specific contributions are noteworthy:

\begin{enumerate}
\item We have adopted a hierarchical image partitioning approach, namely Cuboidal Partitioning of Image Data (CuPID \cite{ahmmed_cupid}), to capture the scene composition structure (SCS) of images. 
\item We have then developed a novel SCS-focused similarity metric, namely the \textit{SCS Similarity Index Measure} (SCSSIM), by exploiting the statistical measures of the hierarchical partitioning of two images.
\end{enumerate}

\section{Scene Composition Structure (SCS)}
\label{sec:scs}

In this section, we first formally define SCS in Section~\ref{sec:scs-def} and then discuss desirable properties of image similarity metrics for comparing SCS in Section~\ref{sec:props}.

\begin{figure}[tb]
    \centering
    \includegraphics[width=0.45\linewidth]{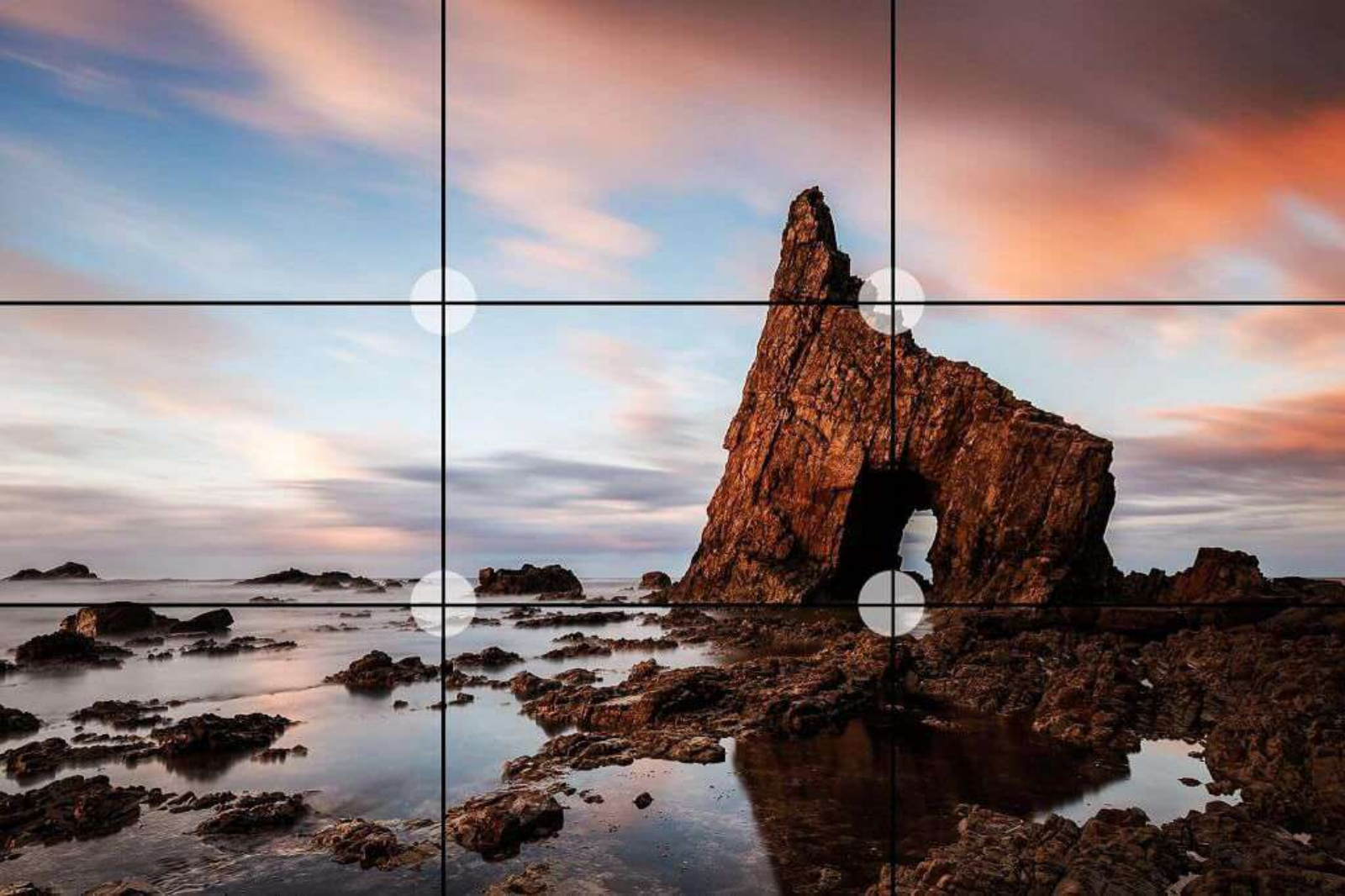}
    \caption{Illustration of the Rule of Thirds grid: Demonstrating a traditional photo composition principle that inspires our approach to Scene Composition Structure (SCS).} \label{fig:rule-of-thirds}
\end{figure}

\subsection{Definition}
\label{sec:scs-def}

SCS can perhaps be understood well from the photo composition rules that are often followed by photographers to improve photo aesthetics. The composition of a photo refers to the way a photographer arranges the visual elements in the scene. The objects in the scene, their lighting, their relative position to the background and the camera, the orientation or zoom level of the camera, all play a role in setting the composition. Changes to any of these will change the composition structure of the scene captured by the photo.

In the world of photography, the \textit{Rule of Thirds}~\cite{koliska_rot_2023} is a longstanding principle to compose aesthetically pleasing images. According to this rule, the visual frame is split into a three-by-three grid with two equally-spaced horizontal and vertical lines as shown in Fig.~\ref{fig:rule-of-thirds}. An aesthetically pleasant photo can then be captured by framing key objects or large geometric structures in the scene along these lines or at their intersections.

\begin{figure}[tb]
    \centering
    \includegraphics[width=\linewidth, trim={90pt, 30pt, 90pt, 20pt}, clip]{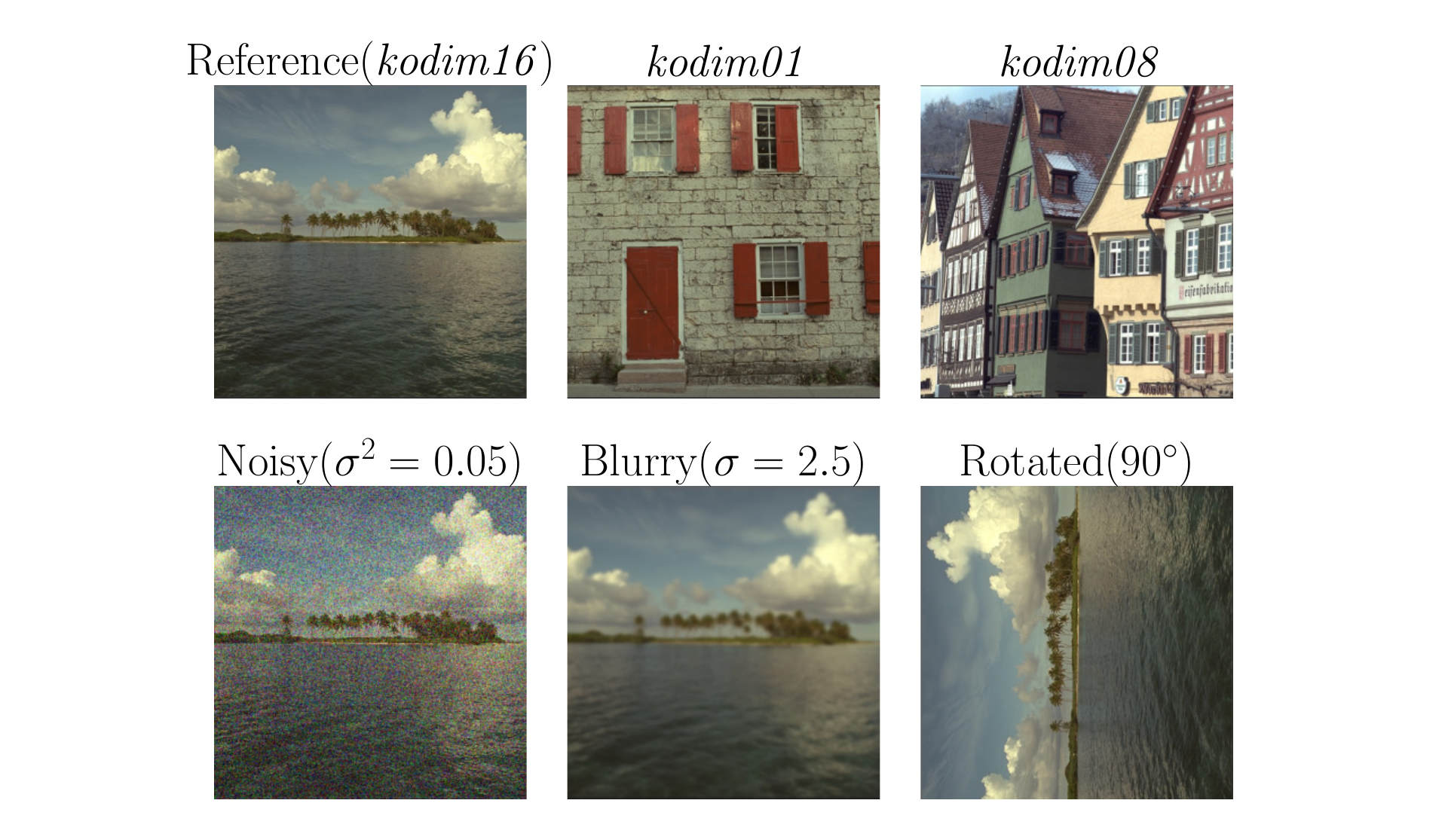}
    \caption{Example  images (top) and distorted versions (bottom) of the reference image (\textit{kodim16}) highlight scenarios where SCS preservation is crucial but often misjudged by conventional metrics.}
    \label{fig:kodims}
\end{figure}

However, not all photographers follow this rule, e.g., the horizon in image \textit{kodim16} (Fig.~\ref{fig:kodims}) is framed at the central line. When images are captured without any human-in-the-loop, e.g., autonomous multimedia sensing and video surveillance, the aesthetics of the images is not always the prime concern. Moreover, many image processing operations, such as cropping and rotating, can also break away from this rule, even though the original images follow the rule.  Even for such images, the presence of ``strong'' horizontal and/or vertical lines can be used as a guide to effectively explain the SCS instead of considering them blindly at equal spaces. For example, the reference image \textit{kodim16} (Fig~\ref{fig:kodims}) has ``strong'' horizontal lines, as the key structures (water, trees, sky) can be split using horizontal lines.

This has motivated us to formally define SCS by explicitly identifying ``strong'' horizontal and vertical lines in an image in a hierarchical fashion so that the earlier lines mimic the fundamental splits of the image, where some statistical optimisations are reached. If such an optimal hierarchical partitioning of an image can be obtained, a similarity metric can then be designed to directly exploit SCS by comparing the statistical measures for each partition of the two images being compared. Recently, CuPID \cite{ahmmed_cupid} has been proposed as a hierarchical image partitioning method, which optimises the sum of squared errors (SSE) measure, with applications in statistical context modeling for efficient image/video data compression. 

\subsection{Desirable Properties of SCS Similarity Metrics}
\label{sec:props}

Consider the images in Fig.~\ref{fig:kodims} from the Kodak dataset \cite{yang2023lossy}. The reference image \textit{kodim16} has ``strong'' horizontal lines. Adding Gaussian noise or blur to this image worsens its quality (human perception), but the SCS remains unaltered. However, the opposite effect is achieved when the image is rotated by $90^\circ$, as its primary SCS lines are now vertical. An ideal SCS similarity metric should evaluate the noisy and blurry images as highly similar and the rotated image as highly dissimilar to the original image.

Image \textit{kodim01} is also somewhat SCS-similar to \textit{kodim16} as the top and bottom floor can be horizontally split. However, image \textit{kodim08} exhibits the opposite, where tall buildings are vertically split. Thus, an ideal similarity metric should evaluate the reference image as appropriately similar to \textit{kodim01} and highly dissimilar to \textit{kodim08}.

Let us now formally define desirable properties of SCS similarity metrics:

\begin{enumerate}
    \item Symmetry: $\mathrm{M}(I_1,I_2)=\mathrm{M}(I_2,I_1)$.
    \item Boundedness: $0\leq\mathrm{M}(I_1,I_2)\leq1$.
    \item Identity: $\mathrm{M}(I_1,I_2)=1$ if $I_1=I_2$.
    \item Invariance: The metric hardly changes with non-compositional distortions.
    \item Monotonicity: The metric decreases monotonically with increase in compositional distortions.
\end{enumerate}

The first three properties are drawn from the necessary properties of any similarity metric. The last two properties are drawn in light of the above discussion.

\begin{table}[tb]
     \caption{Limitations of conventional metrics (highlighted in red) in assessing SCS, contrasted with SCSSIM's ability to correctly assess SCS fidelity, when comparing the reference image \textit{kodim16} with the others in Fig~\ref{fig:kodims}.} \label{tab:metrics}
    \centering
    \addtolength{\tabcolsep}{-3pt}
    \begin{tabular}{lccccc}
         \toprule
         & Noisy & Blurry & Rotated & \textit{kodim01} & \textit{kodim08} \\
         \midrule
         SSIM~\cite{wang_image_2004} & \textcolor{red}{0.09} & \textcolor{red}{0.67} & 0.31 & \textcolor{red}{0.17} & 0.15 \\
         MS-SSIM~\cite{wang_multiscale_2003} & \textcolor{red}{0.45} & 0.89 & 0.27 & \textcolor{red}{0.15} & 0.11 \\
         LPIPS~\cite{zhang_unreasonable_2018} & \textcolor{red}{0.25} & \textcolor{red}{0.48} & 0.29 & 0.24 & 0.22 \\
         CLIP~\cite{hessel_clipscore_2021} & 0.81 & 0.94 & \textcolor{red}{0.93} & 0.70 & \textcolor{red}{0.59} \\
         \textbf{SCSSIM} & \textbf{0.99} & \textbf{0.99} & \textbf{0.09} & \textbf{0.32} & \textbf{0.10} \\
         \bottomrule
    \end{tabular}
\end{table}

\subsection{Existing Similarity Metrics}
\label{sec:existing}

The quality of AI-generated content is often evaluated considering the perception-distortion trade-off \cite{blau_perception-distortion_2018}. Perception or realism is primarily evaluated using deep learning based metrics such as FID~\cite{heusel_gans_2017}, LPIPS~\cite{zhang_unreasonable_2018}, or CLIP Score~\cite{hessel_clipscore_2021}. Whereas, distortion is evaluated using pixel-level analytical metrics such as MSE, PSNR, SSIM~\cite{wang_image_2004} or MS-SSIM~\cite{wang_multiscale_2003}.

Although FID can assess realism, it has little bearing on scene composition. Hence, we evaluate the performance of the remaining SSIM, MS-SSIM, LPIPS\footnote{Although LPIPS is a distance metric, we have converted it into a similarity metric for easier comparison.} (VGG), and CLIP (ViT-L/14) metrics on SCS-similarity in Table~\ref{tab:metrics}. SSIM, MS-SSIM, and LPIPS are heavily based on the human visual system and assess similarity based on human perception. As a result, they penalize noises or artifacts added to images (the noisy and blurry images in Fig.~\ref{fig:kodims}) even when the SCS remains unaffected. CLIP similarity is based on the inherent meaning of two images and hence, hardly affected by changes in SCS, e.g., the $90^\circ$ rotated image in Fig.~\ref{fig:kodims} remains highly similar.

This portrays a fundamental gap in these metrics in capturing SCS, and thus, it is necessary to find a metric that can assess the similarity of two images based on SCS.

\section{Proposed Similarity Metric}
\label{sec:proposed}

In this section, we first briefly introduce CuPID in Section~\ref{sec:cupid} for the sake of completeness and then formally define our novel similarity metric SCSSIM in Section~\ref{sec:SCSSIM}.

\subsection{Cuboidal Partitioning}
\label{sec:cupid}

Let $I$ be an image of width $w$ and height $h$. The sum of squared errors (SSE) $e$ of $I$ is defined by
\begin{equation}
e=\sum_{i \in I}{\|p_i-\mu\|^2}, \label{sse}
\end{equation}
where $p_i$ is the feature-vector of $i$-th pixel and $\mu = \frac{1}{|I|} \sum_{i \in I} p_i$ is the mean feature-vector of $I$. In this paper, the feature vectors are drawn from the RGB colour space. The CuPID algorithm recursively cuts $I$ by finding the ``strongest'' horizontal or vertical lines with optimisation for minimising SSE. 

Initially, the entire image $I$ is considered as a partition, which has $h-1$ horizontal and $w-1$ vertical cuts in total as potential candidates for the ``strongest'' lines in the scene composition. Each cut splits the partition into two [sub] partitions $I_1$ and $I_2$ with SSE $e_1$ and $e_2$, respectively, defined by
\begin{equation}
e_1 = \sum_{i_1 \in I_1}{\|p_{i_1}-\mu_1\|^2} \text{~~~and~~~} e_2 = \sum_{i_2 \in I_2}{\|p_{i_2}-\mu_2\|^2}. \label{sse-12}
\end{equation}
It can be mathematically proven that $e_1 + e_2 \leq e$. Thus, by splitting $I$ into $I_1$ and $I_2$, the total SSE can only decrease. This decrease in the error value can be called the gain $g$ of the cut, defined by
\begin{equation}
g = e - (e_1 + e_2). \label{gain}
\end{equation}
By using greedy optimisation, CuPID selects the cut that maximises the gain, as defined by
\begin{equation}
    \hat{g} = \max_{\forall \text{cuts}} g.
\end{equation}
CuPID recursively continues this process, in a hierarchical fashion, with the [sub] partition of $I$, which can offer the maximum gain next, until a required $N$ number of final partitions are achieved with $N-1$ maximal cuts.

\begin{figure}[tb]
    \centering
    \includegraphics[width=0.8\linewidth, trim=0 0 0 0, clip]{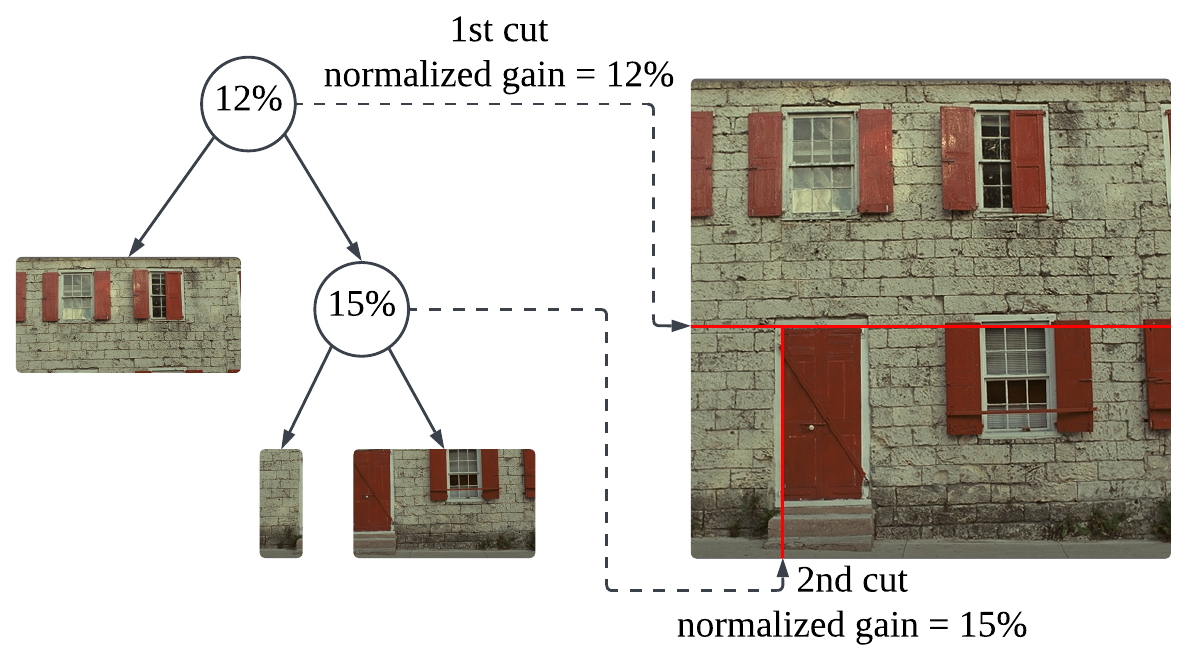}
    \caption{SCS extraction through hierarchical partitioning: Illustrating the initial cuts of the CuPID algorithm and its resulting binary partition tree that forms the foundation for our SCSSIM metric.}
    \label{fig:tree}
\end{figure}

The hierarchical partitioning by CuPID can be represented with a binary tree $T$ (Fig.~\ref{fig:tree}), where each intermediate node represents the cut of the maximum gain with its two children being the newer [sub] partitions. Without any loss of generality, the left and right nodes represent the left and right partitions, respectively, for a vertical cut, or the top and bottom partitions, respectively, for a horizontal cut. The leaf nodes of $T$ are the final partitions of the image.

\subsection{SCSSIM Similarity Metric}
\label{sec:SCSSIM}

Now, we try to build a metric using the CuPID trees of two images being compared for SCS similarity. Naturally, the proposed similarity metric must conform to the desirable properties outlined in Section~\ref{sec:props}.

\begin{figure}[tb]
    \centering
    \includegraphics[width=0.7\linewidth, trim=20 10 10 10, clip]{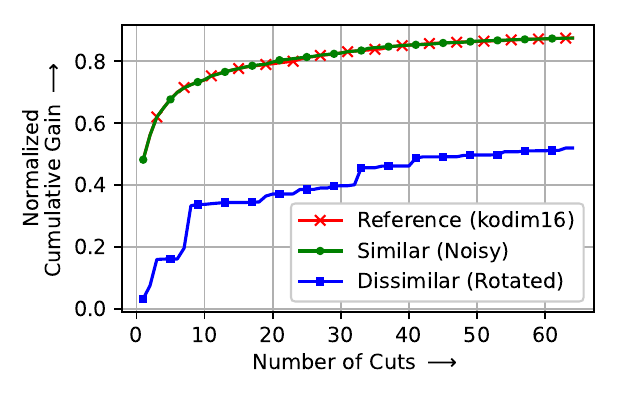}
    \caption{Normalized cumulative gain curves from CuPID trees of similarly composed images are similar in shape and nearly coincide, while curves for dissimilar images have different shapes and are far apart; illustrating how SCSSIM quantifies compositional similarity.}
    \label{fig:comparative-curves}
\end{figure}

The most significant property of each cut in the CuPID tree (Fig.~\ref{fig:tree}) is its gain in SSE. The cumulative sum curve $c$ of these gain values for successive cuts gives us a trend that has high correlations with SCS (Fig.~\ref{fig:comparative-curves}). We argue that this cumulative-gain curve can effectively represent SCS when only the first few $\tilde{N}$ cuts, denoting the ``strongest'' horizontal and vertical lines in the image, are considered. We have used $\tilde{N} = 64$ in all our experiments. As the cumulative sum of SSE gains can vary largely for different images, depending on $e$, the maximum possible cumulative gains, we normalize $c$ in the range $[0,1]$ as follows:
\begin{equation}
c_i = \frac{1}{e}\sum_{j=1}^{i}{\hat{g_j}},  \quad 1 \le i \le \tilde{N}. \label{csumterm}
\end{equation}

In Fig.~\ref{fig:comparative-curves}, cumulative-gain curves are plotted from the CuPID trees of a reference image and two of its distorted versions, one with non-compositional (Noisy) and the other with compositional (Rotated) distortions, all applied on the reference image. We can see that when the tree is derived from a compositionally similar image, the curve is closely similar to the curve produced by the original tree. However, for a compositionally dissimilar image, the curves are quite different in shape and value.

We build on this feature to design our metric. To compare an image $I$ with a reference image $I_0$ we take the CuPID tree $T$ of $\tilde{N}$ cuts for $I_0$. When we apply the cuts of $T_0$ on $I_0$, we get the expected cumulative gain curve for $I_0$. If $I$ is compositionally similar to $I_0$, it should produce a tree $T$ of $\tilde{N}$ cuts that can generate a similar curve when applied on $I_0$. Let the cumulative gain curves of applying $T_0$ and $T$ on $I_0$ be $c_0$ and $c$, respectively. When they are similarly structured, each term of $c$ should be close to those of $c_0$. So, their element-wise division should be fairly close to 1, i.e.,
\begin{equation}
\frac{c_i}{c_{0,i}} \approx 1, \quad 1 \le i \le \tilde{N}. \label{prop-const}
\end{equation}

However, for differently structured curves, these values can range from 0 to $\infty$ in a non-linear fashion. To bring this to a linear range, we apply a logarithmic scale,
\begin{equation}
k_i = \log_e{\frac{c_i}{c_{0,i}}}  = \log_e{c_i} - \log_e{c_{0,i}}, \quad 1 \le i \le \tilde{N}. \label{log}
\end{equation}

We expect $k_i$'s to have values very close to 0 for similar images. So, a good measure of their compositional distance would be their mean $\bar{k} = \frac{1}{\tilde{N}} \sum_{i = 1}^{\tilde{N}} k_i$. However, as $\bar{k}$ is not necessarily bounded, we apply exponential decay on it. Thus, the metric $\mathrm{M_{fr}}(I,I_0)$, which tells us how similar $I$ is to $I_0$ is defined by
\begin{equation}
    \mathrm{M}(I,I_0) = e^{-\lambda \bar{k}^2}, \label{fullref}
\end{equation}
where $\lambda$ is the decay rate. A quadratic decay has been used to avoid negative $\bar{k}$ values and to penalize large differences. Through a sensitivity study for different $\lambda$'s, which is not reported in this paper for page restrictions, $\lambda=5$ is empirically found to give good results, which has been used in the experiments reported in Section~\ref{sec:exp}.

A final symmetric SCSSIM metric can now be defined by averaging $\mathrm{M}$ for both directions as
\begin{equation}
    \mathrm{SCSSIM}(I,I_0) = \frac{1}{2} \Big( \mathrm{M}(I,I_0) + \mathrm{M}(I_0,I) \Big). \label{noref}
\end{equation}
We can tell the metric is bounded and symmetric from the equations of $\mathrm{M}$ and SCSSIM, respectively. Also, it's trivial that, when comparing exact images, SCSSIM will be 1.0.

\section{Experimental Results}
\label{sec:exp}

\begin{figure}[t]
\centering
\includegraphics[width=\linewidth, trim=10 28 10 0, clip]{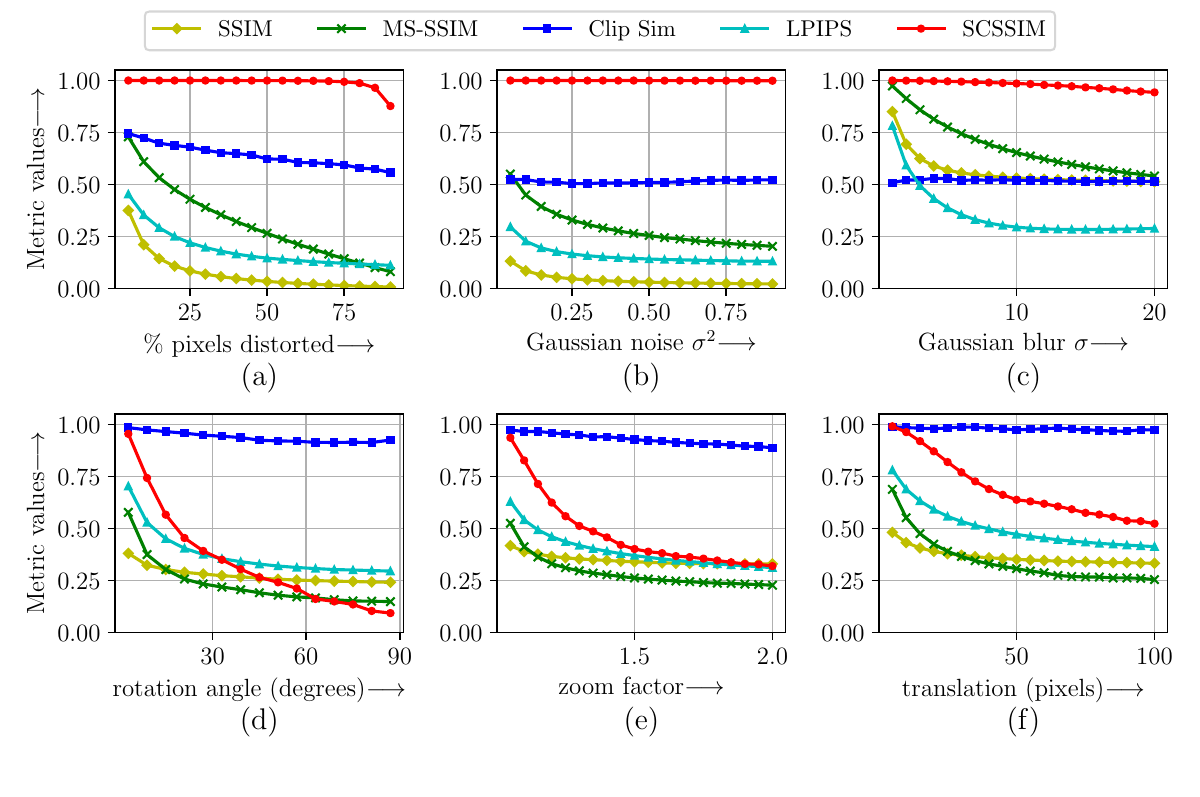}
\caption{Extensive analysis shows SCSSIM exhibits high invariance to non-compositional distortions (a,b,c), reflecting preserved SCS, and a clear monotonic decrease with increasing compositional distortions (d,e,f), reflecting distorted SCS.}
\label{fig:avg_noise}
\end{figure}

We highlight the result for SCSSIM in Table~\ref{tab:metrics} on the images in Fig.~\ref{fig:kodims}. Unlike other similarity metrics, SCSSIM values are consistent with the expectations made in Section~\ref{sec:props}. We have empirically found $\tilde{N} = 64$ to be the most promising. Fewer cuts tend to misrepresent the curved or slanted composition structures, and more cuts start to capture textural details in the image that are not part of SCS. All further results are provided for CuPID trees for this $\tilde{N}$.

\subsection{Effects of Distortions}

Here, we show the invariance and monotonicity properties of SCSSIM by considering how various distortions affect SCS.

To show that SCSSIM is invariant to non-compositional distortions, we consider salt \& pepper noise, Gaussian noise, and Gaussian blur on the full Kodak dataset, as these distortions do not affect the SCS. As evident in Fig.~\ref{fig:avg_noise}(a,b,c), SCSSIM retains a very high value even when the level of distortion is increased significantly. In contrast, SSIM, MS-SSIM, and LPIPS drop values significantly with gradual distortion. The CLIP score also drops with initial distortions and then remains invariant to subsequent changes.

We observe that SCSSIM can correctly ignore local texture and object-level details while comparing the SCS of images (Fig.~\ref{fig:avg_noise}(a,b,c)). However, when too many pixels are distorted, it effectively also distorts the SCS. SCSSIM detects this change by exhibiting lower values for $\geq 80\%$ salt \& pepper noise (Fig.~\ref{fig:avg_noise}(a)) and for $\sigma \geq 10$ blurring (Fig.~\ref{fig:avg_noise}(c)). This property is also seen for Gaussian noise, however, for unrealistically high values of $\sigma^2$ that are outside the range plotted in Fig.~\ref{fig:avg_noise}(b).

In Fig.~\ref{fig:avg_noise}(d,e,f), we consider the effects of camera rotation, zooming, and panning, which inherently change the SCS. In order to perform these distortions without extrapolation, we crop the images at the centre with size $362 \times 362$ for counter-clockwise rotations and $512 \times 512$ pixels for panning to the right. For zooming, only zoom-in operations at the centre are considered with interpolation.

SCSSIM consistently exhibits a clear monotonic decrease with increasing compositional distortion, accurately reflecting the degradation of structural integrity. SSIM, MS-SSIM, and LPIPS also show a similar gradual decrease. In contrast, CLIP's similarity values remain undesirably high and unchanged, failing to capture these critical compositional changes.

\begin{figure}[tb]
\centering
\includegraphics[width=0.8\linewidth, trim= 0 140 0 120,clip]{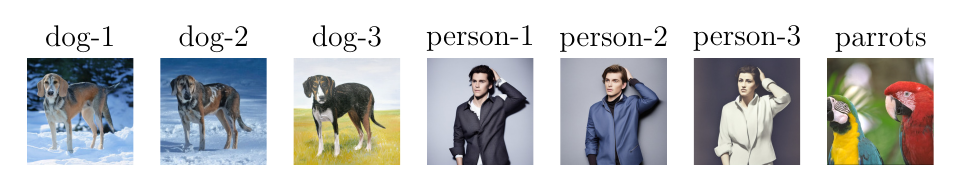}
\includegraphics[width=0.4\linewidth, trim=5 10 10 0,clip]{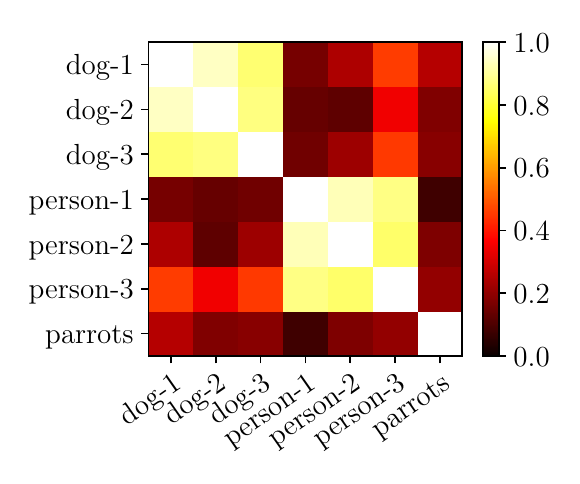}
\caption{This heatmap illustrates SCSSIM's utility in evaluating Generative AI outputs as it accurately groups AI-generated images \cite{zhang_controlnet_2023, bansal_universal_2023} that share the same underlying SCS.}
\label{fig:dog-person-parrot}
\end{figure}

We conclude that while many existing similarity metrics can fairly detect changes in SCS in images, only the proposed SCSSIM metric can fairly detect both SCS similarity and dissimilarity with equal efficiency. This is further evident in Fig.~\ref{fig:dog-person-parrot}, where SCSSIM accurately groups AI-generated images with shared underlying structures. The shared structure was achieved through extensive side information and meticulous prompting. The higher intra-group similarity highlights SCSSIM's potential for assessing and guiding the structural faithfulness of GenAI models.

\begin{figure}[b]
    \centering
    \includegraphics[width=0.7\linewidth, trim=5 15 5 10, clip]{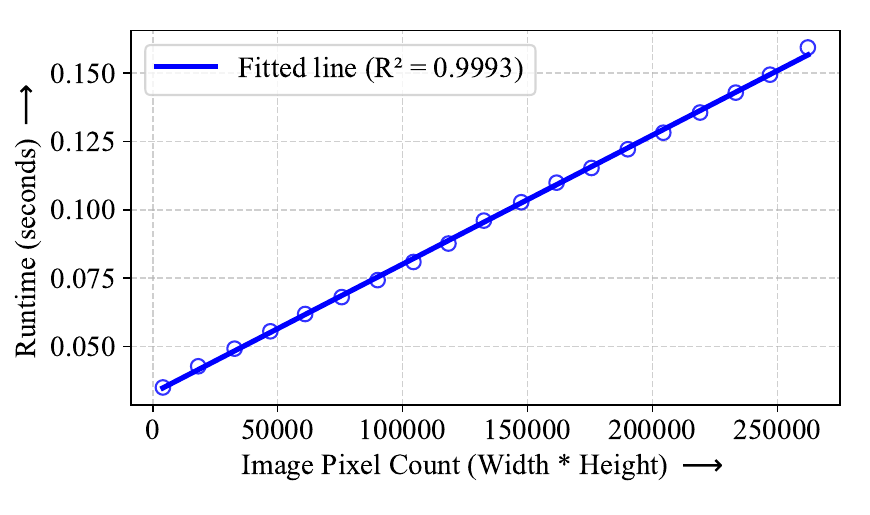}
    \caption{Linear-order Runtime of SCSSIM.}
    \label{fig:runtime}
\end{figure}

\section{Conclusions}
\label{sec:conc}

The proposed SCSSIM is the first similarity metric that can effectively assess whether two images are similar or dissimilar in terms of SCS. Future research directions may include improving AI-generated content using SCSSIM to preserve SCS, which can be further used for learning-based image/video coding.

\newpage

\bibliographystyle{IEEEbib}
\bibliography{refs}

\begin{thebibliography}{10}

\bibitem{rombach_high-resolution_2022}
R.~Rombach, A.~Blattmann, D.~Lorenz, P.~Esser, and B.~Ommer,
\newblock ``High-{Resolution} {Image} {Synthesis} with {Latent} {Diffusion} {Models},''
\newblock in {\em 2022 {IEEE}/{CVF} {Conference} on {Computer} {Vision} and {Pattern} {Recognition} ({CVPR})}, June 2022, pp. 10674--10685.

\bibitem{yang_vcm_2024}
W.~Yang, H.~Huang, Y.~Hu, L.-Y. Duan, and J.~Liu,
\newblock ``Video coding for machines: Compact visual representation compression for intelligent collaborative analytics,''
\newblock {\em IEEE Transactions on Pattern Analysis and Machine Intelligence}, vol. 46, no. 7, pp. 5174--5191, 2024.

\bibitem{rajin_forward_2022}
S.~M. A.~K. Rajin, M.~Murshed, M.~Paul, S.~W. Teng, and J.~Ma,
\newblock ``Human pose based video compression via forward-referencing using deep learning,''
\newblock in {\em 2022 IEEE International Conference on Visual Communications and Image Processing (VCIP)}, 2022, pp. 1--5.

\bibitem{yang_modeling_2022}
Z.~Yang, D.~Liu, C.~Wang, J.~Yang, and D.~Tao,
\newblock ``Modeling image composition for complex scene generation,''
\newblock in {\em Proceedings of the IEEE/CVF Conference on Computer Vision and Pattern Recognition}, 2022, pp. 7764--7773.

\bibitem{zhang_controlnet_2023}
L.~Zhang, A.~Rao, and M.~Agrawala,
\newblock ``Adding conditional control to text-to-image diffusion models,''
\newblock in {\em Proceedings of the IEEE/CVF International Conference on Computer Vision (ICCV)}, October 2023, pp. 3836--3847.

\bibitem{bansal_universal_2023}
A.~Bansal, H.-M. Chu, A.~Schwarzschild, S.~Sengupta, M.~Goldblum, J.~Geiping, and T.~Goldstein,
\newblock ``Universal {Guidance} for {Diffusion} {Models},''
\newblock in {\em 2023 {IEEE}/{CVF} {Conference} on {Computer} {Vision} and {Pattern} {Recognition} {Workshops} ({CVPRW})}, June 2023, pp. 843--852.

\bibitem{wang_image_2004}
Z.~Wang, A.~Bovik, H.~Sheikh, and E.~Simoncelli,
\newblock ``Image quality assessment: from error visibility to structural similarity,''
\newblock {\em IEEE Transactions on Image Processing}, vol. 13, no. 4, pp. 600--612, Apr. 2004.

\bibitem{wang_multiscale_2003}
Z.~Wang, E.~Simoncelli, and A.~Bovik,
\newblock ``Multiscale structural similarity for image quality assessment,''
\newblock in {\em The {Thrity}-{Seventh} {Asilomar} {Conference} on {Signals}, {Systems} \& {Computers}, 2003}, Pacific Grove, CA, USA, 2003, pp. 1398--1402, IEEE.

\bibitem{heusel_gans_2017}
M.~Heusel, H.~Ramsauer, T.~Unterthiner, B.~Nessler, and S.~Hochreiter,
\newblock ``{GANs} {Trained} by a {Two} {Time}-{Scale} {Update} {Rule} {Converge} to a {Local} {Nash} {Equilibrium},''
\newblock in {\em Advances in {Neural} {Information} {Processing} {Systems}}. 2017, vol.~30, Curran Associates, Inc.

\bibitem{zhang_unreasonable_2018}
R.~Zhang, P.~Isola, A.~A. Efros, E.~Shechtman, and O.~Wang,
\newblock ``The {Unreasonable} {Effectiveness} of {Deep} {Features} as a {Perceptual} {Metric},''
\newblock in {\em 2018 {IEEE}/{CVF} {Conference} on {Computer} {Vision} and {Pattern} {Recognition}}, June 2018, pp. 586--595.

\bibitem{hessel_clipscore_2021}
J.~Hessel, A.~Holtzman, M.~Forbes, R.~Le~Bras, and Y.~Choi,
\newblock ``{CLIPScore}: {A} {Reference}-free {Evaluation} {Metric} for {Image} {Captioning},''
\newblock in {\em Proceedings of the 2021 {Conference} on {Empirical} {Methods} in {Natural} {Language} {Processing}}. 2021, pp. 7514--7528, Association for Computational Linguistics.

\bibitem{ahmmed_cupid}
A.~Ahmmed, M.~Murshed, M.~Paul, and D.~Taubman,
\newblock ``A commonality modeling framework for enhanced video coding leveraging on the cuboidal partitioning based representation of frames,''
\newblock {\em IEEE Transactions on Multimedia}, vol. 24, pp. 4446--4457, 2022.

\bibitem{koliska_rot_2023}
M.~Koliska and K.~S.-K. Oh,
\newblock ``Guided by the grid: Raising attention with the rule of thirds,''
\newblock {\em Journalism Practice}, vol. 17, no. 2, pp. 354--373, 2023.

\bibitem{yang2023lossy}
R.~Yang and S.~Mandt,
\newblock ``Lossy image compression with conditional diffusion models,''
\newblock {\em Advances in Neural Information Processing Systems}, vol. 36, pp. 64971--64995, 2023.

\bibitem{blau_perception-distortion_2018}
Y.~Blau and T.~Michaeli,
\newblock ``The {Perception}-{Distortion} {Tradeoff},''
\newblock in {\em 2018 {IEEE}/{CVF} {Conference} on {Computer} {Vision} and {Pattern} {Recognition}}, June 2018, pp. 6228--6237.

\end{thebibliography}

\end{document}